\documentclass[conference]{IEEEtran}
\IEEEoverridecommandlockouts
\usepackage{cite}
\usepackage{amsmath,amssymb,amsfonts}
\usepackage{algorithmic}
\usepackage{graphicx}
\usepackage{textcomp}
\usepackage{xcolor}

\usepackage{amsmath} 
\usepackage{booktabs} 
\usepackage{multirow} 
\usepackage{amssymb} 
\usepackage{subfigure}
\usepackage{graphicx}
\usepackage{diagbox}
\usepackage{threeparttable}
 
\usepackage{stackengine}

\usepackage{color}
\usepackage{bm}

\usepackage{marvosym}

\def\BibTeX{{\rm B\kern-.05em{\sc i\kern-.025em b}\kern-.08em
    T\kern-.1667em\lower.7ex\hbox{E}\kern-.125emX}}
\begin{document}

\title{Surgical Temporal Action-aware Network with Sequence Regularization for Phase Recognition}



\author{\IEEEauthorblockN{Zhen Chen\textsuperscript{1*}, Yuhao Zhai\textsuperscript{2*}, Jun Zhang\textsuperscript{2\Letter}, Jinqiao Wang\textsuperscript{1,3,4,5\Letter}}
\IEEEauthorblockA{\textsuperscript{1}\textit{Centre for Artificial Intelligence and Robotics, Hong Kong Institute of Science\&Innovation, Chinese Academy of Sciences} \\
\textsuperscript{2}\textit{Beijing Friendship Hospital, Capital Medical University} \textsuperscript{3}\textit{Institute of Automation, Chinese Academy of Sciences}\\
\textsuperscript{4}\textit{Wuhan AI Research} \textsuperscript{5}\textit{ObjectEye Inc.}\\
zhen.chen@cair-cas.org.hk, zhaiyuhao@ccmu.edu.cn, zhangjun5986@ccmu.edu.cn, jqwang@nlpr.ia.ac.cn}
\thanks{This work is supported by National Key R\&D Program of China (No. 2022ZD0160601), National Natural Science Foundation of China (No. 62276260, 61976210, 62076235, 62176254, 62006230, 62206283), Beijing Municipal Science \& Technology Commission (No. D17100006517003), and InnoHK program.}
\thanks{Z. Chen and Y. Zhai contribute equally to this work.}
}


\maketitle

\begin{abstract}
To assist surgeons in the operating theatre, surgical phase recognition is critical for developing computer-assisted surgical systems, which requires comprehensive understanding of surgical videos. Although existing studies made great progress, there are still two significant limitations worthy of improvement. First, due to the compromise of resource consumption, frame-wise visual features are extracted by 2D networks and disregard spatial and temporal knowledge of surgical actions, which hinders subsequent inter-frame modeling for phase prediction. Second, these works simply utilize ordinary classification loss with one-hot phase labels to optimize the phase predictions, and cannot fully explore surgical videos under inadequate supervision. To overcome these two limitations, we propose a \textbf{S}urgical \textbf{T}emporal \textbf{A}ction-aware Network with sequence \textbf{R}egularization, named \textbf{STAR-Net}, to recognize surgical phases more accurately from input videos. Specifically, we propose an efficient multi-scale surgical temporal action (MS-STA) module, which integrates visual features with spatial and temporal knowledge of surgical actions at the cost of 2D networks. Moreover, we devise the dual-classifier sequence regularization (DSR) to facilitate the training of STAR-Net by the sequence guidance of an auxiliary classifier with a smaller capacity. Our STAR-Net with MS-STA and DSR can exploit visual features of surgical actions with effective regularization, thereby leading to the superior performance of surgical phase recognition. Extensive experiments on a large-scale gastrectomy surgery dataset and the public Cholec80 benchmark prove that our STAR-Net significantly outperforms state-of-the-arts of surgical phase recognition.
\end{abstract}


\section{Introduction}\label{sec1}
The computer-assisted surgery can improve the quality of interventional healthcare, thereby facilitating patient safety \cite{maier2022surgical,chen2023surgical,Zhai2023}. In particular, surgical phase recognition \cite{garrow2021machine} is significant for developing systems to monitor surgical procedures \cite{panesar2020promises}, schedule surgeons \cite{abdalkareem2021healthcare}, promote surgical team coordination \cite{kennedy2020computer}, and educate junior surgeons \cite{kirubarajan2022artificial}. Compared with offline analysis of surgical videos, online recognition can support decision-making during surgery without using future frames, which is more practical in surgical applications.

Online phase recognition of surgical videos is challenging, and has received great research attention and progress \cite{yi2019hard,zhang2022retrieval,selfdistill}. Earlier works \cite{twinanda2016endonet} formulated this task as the frame-by-frame classification, and used auxiliary annotations of surgical tools for multi-task learning \cite{jin2020multi}. Meanwhile, some works \cite{funke2019_3d_conv,zhang2021swnet,zhang2021surgical} utilized 3D convolutions to capture temporal knowledge of surgical videos. To overcome the huge resource consumption of 3D convolution, mainstream methods \cite{jin2017sv,czempiel2020tecno,czempiel2021opera,jin2021temporal,gao2021trans} first used 2D convolutional neural networks (CNNs) to extract the feature vector of each surgical video frame, and then predicted the surgical phase with the inter-frame temporal relationship aggregated by the long short-term memory (LSTM) \cite{jin2017sv}, temporal convolutions \cite{czempiel2020tecno,farha2019mstcn}, or transformers \cite{czempiel2021opera}. On this basis, recent works \cite{jin2021temporal,gao2021trans} further improved this multi-stage paradigm of phase recognition by leveraging long-range temporal relation among frame-wise feature vectors.

\begin{figure*}[t]
    \centering
    \includegraphics[width=0.90\textwidth]{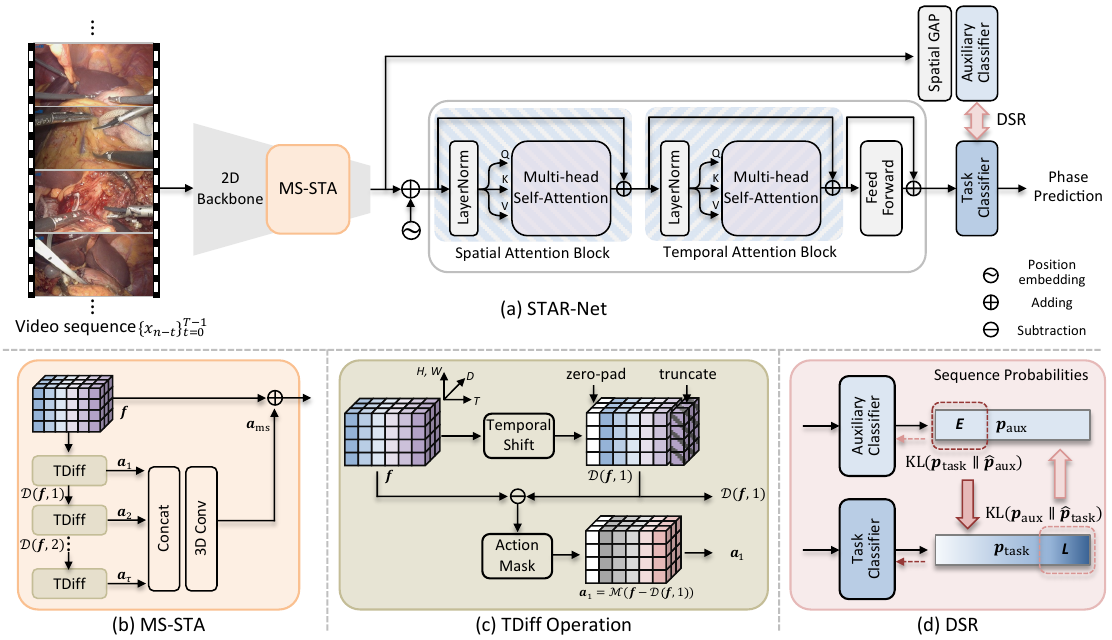}
    \caption{(a) The overview of the STAR-Net, (b) multi-scale surgical temporal action (MS-STA), (c) temporal difference (TDiff) operation, and (d) dual-classifier sequence regularization (DSR). The MS-STA module is inserted into the 2D visual backbone, which progressively conducts TDiff operations to efficiently capture multi-scale surgical action features. The DSR introduces the mutual regularization between the auxiliary classifier and the task classifier at the early and late sequence respectively.}\label{framework}
\end{figure*}

However, existing works \cite{jin2017sv,czempiel2020tecno,czempiel2021opera,jin2021temporal,gao2021trans} on surgical phase recognition suffer from two major limitations, including the insufficient visual information of frame-wise feature vectors, and the inadequate supervision knowledge provided by surgical phase labels. First, most surgical workflow studies \cite{czempiel2020tecno,czempiel2021opera,jin2017sv} first extracted frame-wise feature vectors with 2D networks, and then aggregated these feature vectors for surgical phase prediction. Note that the spatial and temporal information of surgical videos is discarded when 2D networks process frames into feature vectors, thus hindering the subsequent inter-frame modeling. To overcome this bottleneck, we aim to efficiently formulate the surgical actions during feature extraction and provide visual features with spatial and temporal information for sequence modeling and phase prediction. Second, existing works \cite{jin2017sv,czempiel2020tecno,czempiel2021opera,jin2021temporal,gao2021trans} formulated the phase prediction as a classification task of the current frame, and the supervision information provided by the ordinary loss with one-hot phase labels is inadequate, which makes the training susceptible to over-fitting. To guarantee that networks fully learn surgical knowledge as possible, it is beneficial to conduct reasonable regularization in training. Inspired by this idea, we introduce an auxiliary classifier with a smaller capacity to regularize the phase prediction of the input video sequence.

To address these two problems in surgical phase recognition, we propose a \textbf{S}urgical \textbf{T}emporal \textbf{A}ction-aware Network with sequence \textbf{R}egularization, named \textbf{STAR-Net}, from the perspective of feature extraction and surgical supervision. Specifically, we first devise an efficient Multi-Scale Surgical Temporal Action (MS-STA) module and insert it into the visual extraction network, which enables the visual features to perceive the surgical actions at the computational cost of 2D networks. In particular, we progressively conduct Temporal Difference (TDiff) operations to capture multi-scale surgical action features for MS-STA. Moreover, we devise the Dual-classifier Sequence Regularization (DSR) to regularize the training of STAR-Net by introducing an auxiliary classifier with a smaller capacity. As such, this auxiliary classifier regularizes the task classifier at the early sequence to prevent over-fitting, and the task classifier provided with spatial and temporal knowledge enhances the auxiliary classifier at the late sequence in turn. With the proposed MS-STA and DSR, our STAR-Net can exploit visual features with the knowledge of surgical actions and learn from abundant surgical supervision, thereby leading to the superior performance of surgical phase recognition. We perform extensive experiments on a large-scale gastrectomy surgery dataset and the public Cholec80 benchmark to validate the effectiveness of our STAR-Net, which outperforms state-of-the-art surgical phase recognition methods by a large margin.

\section{Methodology}
\subsection{Overview}
As illustrated in Fig.~\ref{framework} (a), our STAR-Net predicts the phase of each frame in surgical videos to achieve online phase recognition. Following previous studies \cite{jin2021temporal}, our STAR-Net classifies the current frame $\boldsymbol{x}_{n}$ as one of $C$ surgical phases by taking the current frame and $T-1$ preceding frames as sequence input $\{\boldsymbol{x}_{n-t}\}_{t=0}^{T-1}$. By progressively shifting the sequence input over time, the STAR-Net can predict the surgical phase of each frame in the entire video.

Specifically, the STAR-Net first utilizes a 2D CNN with the MS-STA module as the backbone to extract visual features with spatial and temporal information of surgical actions. Then, a transformer with spatial and temporal attention blocks efficiently aggregates visual features by exploiting global relationships in spatial and temporal dimensions sequentially. Finally, we introduce the DSR with an auxiliary classifier to mutually regularize sequence predictions produced by the task classifier, thereby facilitating the training of the STAR-Net.

\subsection{Multi-Scale Surgical Temporal Action for Visual Features}
Existing studies \cite{czempiel2020tecno,czempiel2021opera} extracted frame-wise visual information into feature vectors, which lost spatial and temporal information of surgical videos. As a result, the surgical actions in surgical videos are not well represented, thereby leading to inaccurate modeling of the inter-frame relation. To address this problem, we propose the MS-STA module to efficiently model the multi-scale surgical temporal actions during visual extraction of the 2D backbone, which provides visual features with spatial and temporal knowledge for STAR-Net. 

As shown in Fig.~\ref{framework} (b), the MS-STA integrates visual features $\boldsymbol{f}\in \mathbb{R}^{T\times H\times W\times D}$ of video sequences with multi-scale temporal information of surgical actions to facilitate surgical phase recognition, where $T$ is the length of the input sequence, and $H$, $W$ and $D$ are the numbers of height, width and channel dimensions of visual features. In particular, we devise the Temporal Difference (TDiff) operation to capture surgical actions between two adjacent frames, which can be used for longer range surgical actions based on previous operations progressively. 
In Fig.~\ref{framework} (c), the input visual features $\boldsymbol{f}$ of TDiff operation are first shifted along the temporal dimension for one frame as delayed features $\mathcal{D}(\boldsymbol{f},1)$, where the first and the last frame is performed with zero-padding and truncation, respectively. Then, we subtract the delayed features $\mathcal{D}(\boldsymbol{f},1)$ from the input visual features $\boldsymbol{f}$ element-wise to calculate the surgical action features of each frame relative to the previous adjacent frame, as follows:
\begin{equation}
    \boldsymbol{a}_{1}=\mathcal{M}(\boldsymbol{f}-\mathcal{D}(\boldsymbol{f},1)),
\end{equation}
where action mask $\mathcal{M}(\cdot)$ sets the first frame substraction to $0$. Note that the TDiff operation efficiently captures surgical action features for each frame with only one shift operation and element-wise subtraction. In this way, we obtain surgical action features $\boldsymbol{a}_{1}$ and delayed features $\mathcal{D}(\boldsymbol{f},1)$ as the output of TDiff operation.

With the delayed features, the MS-STA can further perform the TDiff operation to progressively generate the action features with a longer temporal range, \textit{e.g.}, $\mathcal{D}(\boldsymbol{f},2)=\mathcal{D}(\mathcal{D}(\boldsymbol{f},1),1)$. By conducting multiple TDiff operations sequentially in Fig.~\ref{framework} (b), we concatenate these surgical action features $\{\boldsymbol{a}_{k}\}_{k=1}^{\tau}$ with multiple temporal scales, where $[\boldsymbol{a}_{1}, \boldsymbol{a}_{2}, \cdots,\boldsymbol{a}_{\tau}]\in \mathbb{R}^{T\times \tau\times H\times W\times D}$ and $\tau$ denotes the number of temporal scales, and then perform a 3D convolution to integrate the multi-scale temporal features of surgical actions, as follows:
\begin{equation}
    \boldsymbol{a}_{\rm ms}=\boldsymbol{W}\circledast [\boldsymbol{a}_{1}, \boldsymbol{a}_{2}, \cdots,\boldsymbol{a}_{\tau}],
\end{equation}
where $\boldsymbol{a}_{\rm ms}\in \mathbb{R}^{T\times H\times W\times D}$, $\boldsymbol{W}$ is the parameters of a 3D convolutional layer and $\circledast$ is the convolution operation. In contrast to the burdensome 3D convolutional networks, we only insert one 3D convolutional layer into the STAR-Net to integrate multi-scale temporal features of surgical actions, which perceive the surgical actions at the computational cost of 2D networks.

Finally, we add the multi-scale surgical action features $\boldsymbol{a}_{\rm ms}$ with the input features $\boldsymbol{f}$ as residual learning, which can provide each frame with the knowledge of surgical actions for the surgical phase recognition. Different from TSM \cite{lin2019tsm} that shifted partial channels for temporal information at different layers, our MS-STA can efficiently capture multi-scale temporal information of surgical actions at once, while preserving the channel alignment of visual features, thereby providing surgical action features for phase recognition.

\begin{figure}[t]
    \centering
    \includegraphics[width=0.45\textwidth]{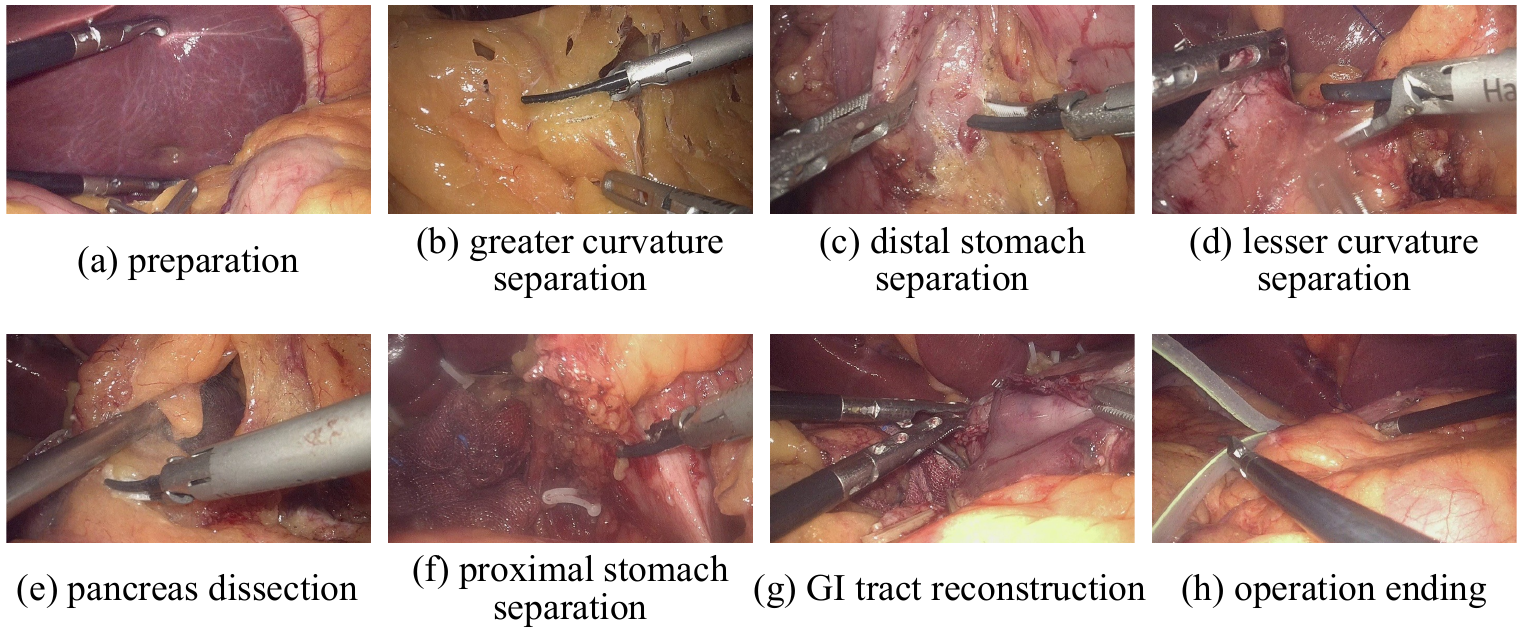}
    \caption{Typical examples of eight phases in gastrectomy phase dataset. Each surgical phase carries a distinct and specific clinical significance and serves as the necessary procedure of the gastrectomy.
    }
    \label{fig_phase}
\end{figure}

\subsection{Dual-Classifier Sequence Regularization}
With multi-scale surgical action features provided by MS-STA, the STAR-Net can predict the surgical phase with discriminative spatial and temporal features. However, existing works \cite{twinanda2016endonet,czempiel2020tecno,jin2021temporal,gao2021trans} employed ordinary classification loss, \textit{e.g.}, the cross-entropy loss and its variants, to train the network, which cannot provide sufficient supervision for the training. Since the phase label $\boldsymbol{y}$ is a one-hot vector to indicate the correct class, the cross-entropy loss $L_{\rm CE}=-\sum_{c=1}^{C}\boldsymbol{y}_{c}\log \boldsymbol{p}_{c}$ merely produces a single non-zero constraint among these $C$ terms to supervise the network training. As a result, the lack of supervision makes the network prone to over-fitting, and thus restricts the performance of surgical phase recognition.

To address the lack of supervision, we devise the Dual-classifier Sequence Regularization (DSR) to regularize sequence predictions by introducing a frame-wise auxiliary classifier, as illustrated in Fig.~\ref{framework} (d). With the tokens of each frame provided by the transformer in STAR-Net, the task classifier can generate frame-wise phase predictions of the input video sequence, where the predicted probabilities are denoted as $\boldsymbol{p}_{\rm task}$. Meanwhile, the auxiliary classifier uses the sequence features extracted by the 2D visual backbone, and performs spatial global average pooling to predict the phase probabilities $\boldsymbol{p}_{\rm aux}$ of each frame.

\begin{table*}[t]
\centering
\caption{Comparison with state-of-the-arts on gastrectomy phase dataset. Best and second best results are {\bf highlighted} and \underline{underlined}.}\label{sota_compare_gas}
\begin{tabular}{c|cccc|c}
    \toprule[1pt]
    {Method} & AC (\%) & PR (\%) & RE (\%) & JA (\%) & P-value \\ \hline
    PhaseNet \cite{twinanda2016endonet} & 72.9$_{\pm 7.2}$ & 66.5$_{\pm 17.6}$ & 70.4$_{\pm 5.1}$ & 52.2$_{\pm 13.6}$ & 2.8$\times10^{-17}$ \\
    SV-RCNet \cite{jin2017sv} & 84.3$_{\pm 7.6}$ & 79.8$_{\pm 9.4}$ & 78.9$_{\pm 7.9}$ & 66.1$_{\pm 10.6}$ & 1.4$\times10^{-12}$ \\
    TeCNO \cite{czempiel2020tecno} & 85.4$_{\pm 7.1}$ & 80.9$_{\pm 9.5}$ & 80.3$_{\pm 7.7}$ & 68.0$_{\pm 10.9}$ & 1.3$\times10^{-9}$  \\
    TMRNet \cite{jin2021temporal} & 86.8$_{\pm 6.2}$ & 85.1$_{\pm 7.1}$ & 81.9$_{\pm 8.3}$ & 71.8$_{\pm 9.3}$ & 2.7$\times10^{-7}$ \\
    Trans-SVNet \cite{gao2021trans} & 87.7$_{\pm 6.0}$ & 85.1$_{\pm 6.7}$ & 82.0$_{\pm 8.4}$ & 71.9$_{\pm 9.4}$ & 2.5$\times10^{-6}$ \\ \hline
    \multicolumn{1}{l|}{STAR-Net \textit{w/o} MS-STA, DSR} & 85.1$_{\pm 6.3}$ & 80.5$_{\pm 11.6}$ & 81.5$_{\pm 5.0}$ & 68.6$_{\pm 10.8}$ & 1.6$\times10^{-8}$ \\
    \multicolumn{1}{l|}{STAR-Net \textit{w/o} MS-STA} & 86.8$_{\pm 6.2}$ & 81.8$_{\pm 10.3}$ & \underline{82.7}$_{\pm 6.1}$ & 69.6$_{\pm 10.2}$ & 2.6$\times10^{-8}$ \\
    \multicolumn{1}{l|}{STAR-Net \textit{w/o} DSR} & \underline{87.9}$_{\pm 6.9}$ & \underline{85.5}$_{\pm 7.1}$ & 82.3$_{\pm 8.5}$ & \underline{72.0}$_{\pm 9.1}$ & 3.0$\times10^{-6}$ \\
    STAR-Net & {\bf 89.2}$_{\pm 6.1}$ & {\bf 86.6}$_{\pm 6.4}$ & {\bf 83.7}$_{\pm 8.1}$ & {\bf 73.5}$_{\pm 9.0}$ & - \\
    \bottomrule[1pt]
\end{tabular}
\end{table*}

Since MS-STA provides multi-scale temporal information of surgical actions, the auxiliary classifier can achieve relatively satisfactory prediction for each video frame. Considering that the small number of previous frames in the early sequences $\boldsymbol{E}$ cannot provide sufficient temporal knowledge for the task classifier after the transformer, we adopt the auxiliary classifier with a smaller capacity to regularize the predicted probabilities $\boldsymbol{p}_{\rm task}$ of the task classifier. This provides effective regularization for the training of STAR-Net, thereby avoiding over-fitting. On the other hand, due to the lack of long-range surgical video knowledge, the auxiliary classifier is inferior to the task classifier on the late sequences $\boldsymbol{L}$, and thus we further improve the auxiliary classifier with the task classifier. In turn, this can promote the learning of the task classifier with an improved auxiliary classifier. Therefore, the objective of our DSR is summarized as follows:
\begin{equation}\label{eq_dsr}
    L_{\rm DSR}=\sum\nolimits_{i\in \boldsymbol{E}}{\rm KL}(\boldsymbol{p}^{(i)}_{\rm task}||\hat{\boldsymbol{p}}^{(i)}_{\rm aux}) + \sum\nolimits_{j\in \boldsymbol{L}}{\rm KL}(\boldsymbol{p}^{(j)}_{\rm aux}||\hat{\boldsymbol{p}}^{(j)}_{\rm task}),
\end{equation}
where ${\rm KL}$ is the Kullback–Leibler divergence to measure the distance between two probabilities, and $\hat{\boldsymbol{p}}$ represents stopping the gradients from $\boldsymbol{p}$ by regarding it as constants. Therefore, the first term in Eq.~\eqref{eq_dsr} optimizes ${\boldsymbol{p}_{\rm task}}$ on the early sequences $\boldsymbol{E}$, while the second term optimizes ${\boldsymbol{p}_{\rm aux}}$ on the late sequences $\boldsymbol{L}$. In this way, the DSR can facilitate the training of STAR-Net with the sequence regularization between the task classifier and the auxiliary classifier.

\begin{figure}[t]  
    \centering
    \includegraphics[width=0.37\textwidth]{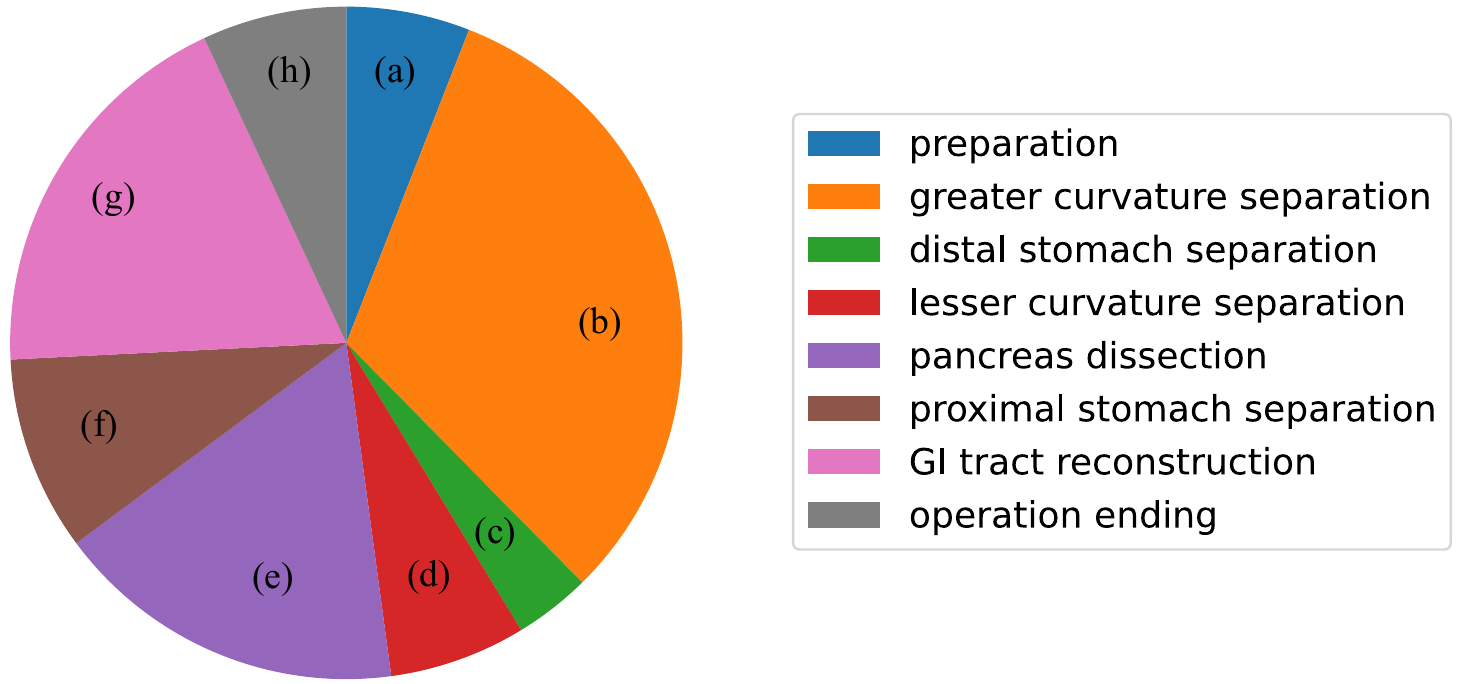}
    \caption{The proportion of eight phases in gastrectomy phase dataset. The inherent imbalance of surgical phases makes online recognition challenging.}
    \label{fig_phase_pie}  
\end{figure}  

\subsection{Training and Inference}
Following the efficient multi-stage training paradigm in previous works \cite{jin2021temporal,gao2021trans}, we first train the 2D visual backbone with MS-STA using the cross-entropy loss $L_{\rm CE}$, and generate frame features with spatial and temporal knowledge. Then, we train the transformer with the task and auxiliary classifiers under DSR for surgical phase recognition, as follows:
\begin{equation}
    L = L_{\rm CE}+\lambda L_{\rm DSR},
\end{equation}
where the coefficient $\lambda$ controls the trade-off between $L_{\rm DSR}$ and the cross-entropy loss $L_{\rm CE}$ of phase predictions. In the inference, the well-trained STAR-Net sequentially conducts the 2D visual backbone with MS-STA and the transformer with spatial and temporal attention blocks to extract visual features, and performs the online frame-wise prediction using the task classifier for the surgical video streaming in an end-to-end manner.

\section{Experiment}
\subsection{Dataset and Implementation Details}
\subsubsection{Gastrectomy Phase Dataset}
To evaluate the online phase recognition of surgical videos, we collect a large-scale laparoscopic gastrectomy dataset consisting of $100$ surgical videos from different gastric cancer patients, and its data size is $22.1$ times\footnote{The size of the dataset is measured in the number of pixels.} of the Cholec80 dataset \cite{twinanda2016endonet}. The surgical videos are recorded with $1,920\times 1,080$ resolution and $25$ frame-per-second (fps). The average length of surgical videos is $2.53$ hours.  All surgical videos are annotated by two surgeons with expertise in gastric cancer surgery. Each frame of surgical videos is assigned to one out of eight surgical phases, including the preparation, the greater curvature separation, the distal stomach separation, the lesser curvature separation, the pancreas dissection, the proximal stomach separation, the gastrointestinal (GI) tract reconstruction, and the operation ending. We randomly split the dataset at the patient-level, as $70$ videos for training and $30$ videos for test.


To elaborate the collected gastrectomy phase dataset for surgical phase recognition, we show typical examples of eight phases in the gastrectomy surgery in Fig.~\ref{fig_phase}. It is evident that each of these surgical phases carries distinct and specific clinical significance, and together these phases constitute the entire procedures of gastrectomy. Moreover, the proportion of eight phases is illustrated in Fig.~\ref{fig_phase_pie}. It is worth noting that the inherent imbalance of these eight phases makes it more difficult to accurately achieve the online phase recognition.

\begin{figure}[t]
    \centering
    \includegraphics[width=0.47\textwidth]{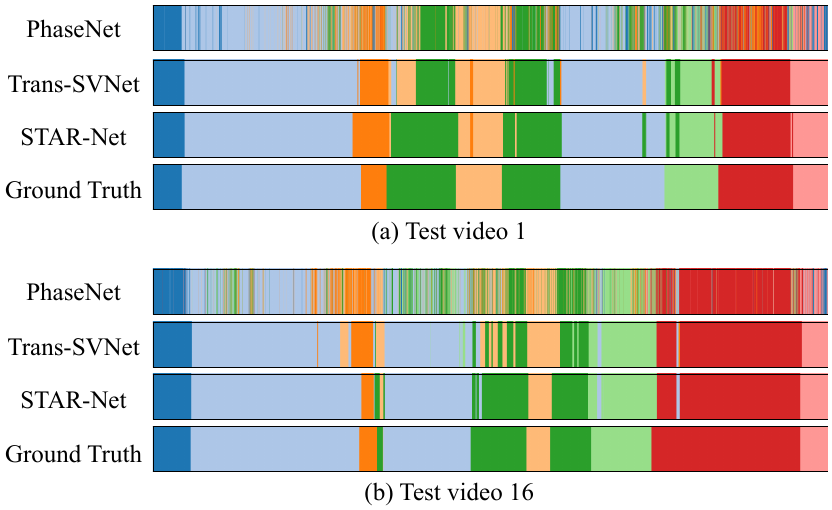}
    \caption{Color-coded ribbon comparison of PhaseNet, Trans-SVNet, STAR-Net and ground truth.}
    \label{fig_bar_plot}
\end{figure}

\begin{table*}[th]
\centering
	\caption{Comparison with state-of-the-arts on Cholec80 dataset. Best and second best results are {\bf highlighted} and \underline{underlined}.}\label{sota_compare_cholec80}
\begin{tabular}{c|cccc|cc}
    \toprule[1pt]
    {Method} & AC (\%) & PR (\%) & RE (\%) & JA (\%) & Param {\tiny ($10^{7}$)} & FLOPs {\tiny ($10^{10}$)}\\ \hline
    PhaseNet \cite{twinanda2016endonet} & 78.8$_{\pm 4.7}$ & 71.3$_{\pm 15.6}$ & 76.6$_{\pm 16.6}$ & - & 4.23 & {\bf 0.07} \\
    SV-RCNet \cite{jin2017sv} & 85.3$_{\pm 7.3}$ & 80.7$_{\pm 7.0}$ & 83.5$_{\pm 7.5}$ & - & 2.88 & 4.14 \\
    UATD \cite{ding2023less} & 88.6$_{\pm 6.7}$ & 86.1$_{\pm 6.7}$ & 88.0$_{\pm 10.1}$ & 73.7$_{\pm 10.2}$ & 2.80 & 5.72 \\
    TeCNO \cite{czempiel2020tecno} & 88.6$_{\pm 7.8}$ & 86.5$_{\pm 7.0}$ & 87.6$_{\pm 6.7}$ & 75.1$_{\pm 6.9}$ & 2.36 & 8.29 \\
    MTRCNet-CL \cite{jin2020multi} & 89.2$_{\pm 7.6}$ & 86.9$_{\pm 4.3}$ & 88.0$_{\pm 6.9}$ & - & 2.98 & 4.14 \\
    TMRNet \cite{jin2021temporal} & 89.2$_{\pm 9.4}$ & 89.7$_{\pm 3.5}$ & {\bf 89.5}$_{\pm 4.8}$ & 78.9$_{\pm 5.8}$ & 6.30 & 24.86 \\
    Trans-SVNet \cite{gao2021trans} & \underline{90.3}$_{\pm 7.1}$ & \underline{90.7}$_{\pm 5.0}$ & 88.8$_{\pm 7.4}$ & \underline{79.3}$_{\pm 6.6}$ & \underline{2.37} & 12.47\\
    STAR-Net & {\bf 91.2}$_{\pm 5.3}$ & {\bf 91.6}$_{\pm 3.4}$ & \underline{89.2}$_{\pm 9.4}$ & {\bf 79.5}$_{\pm 8.1}$ & {\bf 1.68} & \underline{3.92}\\
    \bottomrule[1pt]
\end{tabular}
\end{table*}

\begin{figure*}[t]
    \centering
    \includegraphics[width=0.95\textwidth]{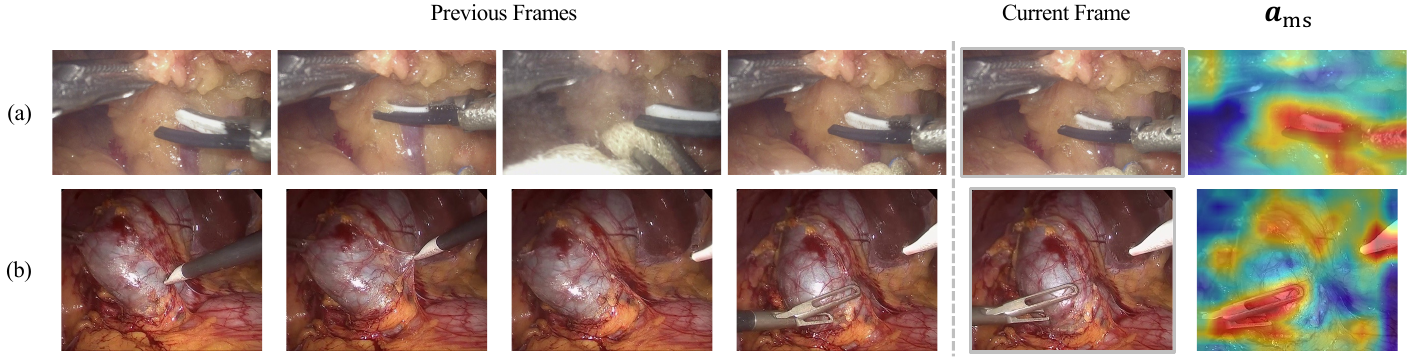}
    \caption{Visualization of surgical action features $\boldsymbol{a}_{\rm ms}$ of MS-STA in (a) gastrectomy and (b) Cholec80 datasets. The motion of the ultrasound knife, grasper and hook is captured in MS-STA, which provides spatial and temporal information for phase recognition.}
    \label{surgical_heatmap}
\end{figure*}

\subsubsection{Cholec80 Dataset} 
We further perform the comparison on public Cholec80 dataset \cite{twinanda2016endonet} of laparoscopic cholecystectomy procedures, which contains $80$ surgical videos with a resolution of $854\times 480$ or $1,920\times 1,080$ at $25$ fps. The surgery procedures are divided into seven surgical phases, including the preparation, the calot triangle dissection, the clipping and cutting, the gallbladder dissection, the gallbladder packaging, the cleaning and coagulation, and the gallbladder retraction. We exactly follow the standard splits \cite{twinanda2016endonet,jin2021temporal}, \textit{i.e.}, the first $40$ videos for training and the rest $40$ videos for test.

\subsubsection{Implementation Details}
We compare STAR-Net with state-of-the-arts using PyTorch \cite{pytorch} on a single NVIDIA A100 GPU. In our STAR-Net, we adopt ResNet-18 \cite{resnet} as the 2D visual backbone for feature extraction, and implement the temporal attention block with causal mask \cite{czempiel2021opera} to achieve online recognition without using future frames. For MS-STA, the temporal scale $\tau$ is set as $5$, and the sequence length $T$ is $20$. The coefficient $\lambda$ of $L_{\rm DSR}$ is set as $1.0$, and $\boldsymbol{E}$ and $\boldsymbol{L}$ are set as the $20\%-60\%$ and $80\%-100\%$ ranges of input video sequences, respectively. All models are optimized in SGD with the batch size of $32$. The learning rate is initialized as $1\times 10^{-3}$ and halved after every $5$ epochs.

\subsubsection{Evaluation Metrics}

We adopt four commonly-used metrics to comprehensively evaluate the performance of surgical phase recognition, including accuracy (AC), precision (PR), recall (RE) and Jaccard (JA). Following the evaluation protocol in previous works \cite{twinanda2016endonet,jin2021temporal}, we calculate PR, RE and JA in the phase-wise manner, and report the average and standard deviation. The AC represents the percentage of frames correctly classified into ground truth. To perform fair comparisons, the selected state-of-the-art methods are evaluated with the same criteria as the STAR-Net. Note that all experiments are performed in the online mode, where future information is not accessible when estimating the current frame.

\subsection{Comparison on Gastrectomy Dataset}

\subsubsection{Comparison with state-of-the-arts} 

To verify the effectiveness of our STAR-Net, we perform a comprehensive comparison with the state-of-the-art methods \cite{twinanda2016endonet,jin2017sv,czempiel2020tecno,jin2021temporal,gao2021trans}. As illustrated in Table \ref{sota_compare_gas}, our STAR-Net achieves the best performance among these methods, with the AC of $89.2\%$ and JA of $73.5\%$. Noticeably, our STAR-Net outperforms the transformer-based method, Trans-SVNet~\cite{gao2021trans}, by a large margin, \textit{e.g.}, $1.5\%$ in AC and $1.6\%$ in JA. In addition, we conduct the t-test of AC among paired test videos, which confirms a significant advantage of our STAR-Net over \cite{twinanda2016endonet,jin2017sv,czempiel2020tecno,jin2021temporal,gao2021trans} with P-values $<1\times 10^{-5}$. These results demonstrate the performance advantage of our STAR-Net over state-of-the-arts on gastrectomy phase recognition.

\subsubsection{Ablation Study} 
As elaborated in Table \ref{sota_compare_gas}, we perform the detailed ablation study to validate the effectiveness, by implementing three ablative baselines of STAR-Net without MS-STA or DSR. Compared with the baseline without both MS-STA and DSR, the MS-STA can bring an improvement of $2.8\%$ in AC and $3.4\%$ in JA, which reveals the impact of surgical actions on the task. Meanwhile, the DSR can also increase the baseline with $1.7\%$ in AC, which validates the sequence regularization of the auxiliary classifier benefits the training of STAR-Net. The ablation experiments indicate that the proposed MS-STA and DSR are crucial to improving the performance of STAR-Net on surgical phase recognition. 

\subsubsection{Qualitative Results of Phase Recognition} 
We further qualitatively compare our STAR-Net with Trans-SVNet \cite{gao2021trans} and PhaseNet \cite{twinanda2016endonet} by presenting the color-coded ribbon results on gastrectomy dataset. As shown in Fig.~\ref{fig_bar_plot}, our STAR-Net outperforms both PhaseNet \cite{twinanda2016endonet} and Trans-SVNet \cite{gao2021trans}, and is the closest to ground truth. In this way, these qualitative results confirm the superiority of our STAR-Net in surgical video analysis.

\subsection{Comparison on Cholec80 Dataset}
To further evaluate the performance of phase recognition, we perform the comparison with more state-of-the-arts \cite{jin2020multi,ding2023less} on the public Cholec80 benchmark in terms of performance and efficiency. In Table \ref{sota_compare_cholec80}, our STAR-Net achieves the overwhelming performance with the best AC of $91.2\%$, PR of $91.6\%$ and JA of $79.5\%$. Furthermore, our STAR-Net demonstrates superior efficiency in comparison to existing algorithms \cite{jin2017sv,ding2023less,czempiel2020tecno,jin2020multi,jin2021temporal,gao2021trans} with the minimal parameters and computation except for the frame-wise 2D CNN \cite{twinanda2016endonet}. These competitive experimental results confirm the superiority of our STAR-Net on surgical phase recognition.

\subsection{Qualitative Analysis of Surgical Temporal Action}
To analyze the surgical temporal action, we further visualize the multi-scale action features $\boldsymbol{a}_{\rm ms}$ of MS-STA, as shown in Fig.~\ref{surgical_heatmap}. Compared with the current frame, the MS-STA can accurately capture the surgical actions from several previous frames, where multi-scale action features $\boldsymbol{a}_{\rm ms}$ highlight the instrument motions on gastrectomy and Cholec80 datasets. For example, the motion of the ultrasound knife, grasper and hook is perceived by the multi-scale action features of MS-STA in Fig.~\ref{surgical_heatmap}. In this way, the MS-STA provides visual features with the spatial and temporal information of surgical actions for the STAR-Net, thereby facilitating the phase recognition tasks.

\section{Conclusion}
In this work, we propose the STAR-Net to promote online surgical phase recognition efficiently. Specifically, we first devise the MS-STA module to integrate the visual features with the multi-scale temporal knowledge of surgical actions, which enables the STAR-Net to process the surgical video sequence with more abundant surgical information. Moreover, we introduce the DSR to regularize the training of STAR-Net over the frame prediction of video sequences using an auxiliary classifier. Extensive experiments on gastrectomy and cholecystectomy surgical datasets confirm the remarkable advantages of our STAR-Net over state-of-the-art works in terms of performance and efficiency, as well as the perception of surgical temporal actions. 
%



\end{document}